\documentclass{article}

\PassOptionsToPackage{sort, authoryear, round}{natbib}
\usepackage[final]{neurips_2021}

\usepackage[utf8]{inputenc} % allow utf-8 input
\usepackage[T1]{fontenc}    % use 8-bit T1 fonts
\usepackage{hyperref}       % hyperlinks
\usepackage{url}            % simple URL typesetting
\usepackage{booktabs}       % professional-quality tables
\usepackage{amsfonts}       % blackboard math symbols
\usepackage{nicefrac}       % compact symbols for 1/2, etc.
\usepackage{microtype}      % microtypography
\usepackage{xcolor}         % colors
\usepackage{graphicx}

\usepackage[bottom]{footmisc}

\title{Who Decides if AI is Fair? The Labels Problem in Algorithmic Auditing}

\author{Abhilash ~Mishra\thanks{Both authors contributed equally.}~~\thanks{Corresponding author}\\
  The University of Chicago \& EquiTech Futures\\
  \texttt{abhilash@equitechfutures.com} \\
 \And
Yash ~Gorana\footnotemark[1] \\
  EquiTech Futures  \\
  \texttt{yash@equitechfutures.com} \\  % examples of more authors

}

\begin{document}

\maketitle

\begin{abstract}
Labelled "ground truth" datasets are routinely used to evaluate and audit AI algorithms applied in high-stakes settings. However, there do not exist widely accepted benchmarks for the quality of labels in these datasets. We provide empirical evidence that quality of labels can significantly distort the results of algorithmic audits in real-world settings. Using data annotators typically hired by AI firms in India, we show that fidelity of the ground truth data can lead to spurious differences in performance of ASRs between urban and rural populations. After a rigorous, albeit expensive, label cleaning process, these disparities between groups disappear. Our findings highlight how trade-offs between label quality and data annotation costs can complicate algorithmic audits in practice.  They also emphasize the need for development of consensus-driven, widely accepted benchmarks for label quality. 

%is widely recognized as critical for ensuring fairness and equity. One strategy for algorithmic auditing is to test the algorithm against a benchmark labelled dataset built by auditors.

\end{abstract}

\section{Introduction}
There is now a compelling body of evidence that AI systems deployed in high-stakes domains like healthcare, criminal justice, and education can be biased against socioeconomically disadvantaged communities. Examples include bias in facial and object recognition \citep{buolamwini2018gender, devries2019recognition}, speech \citep{koenecke2020racial}, and healthcare systems \citep{obermeyer2019dissecting}. These concerns have been heightened with the growing dominance of so-called foundation models \citep{bommasani2021risks} which can amplify biases across a wide range of applications through single points of failure. 

Fairness and ethical concerns around AI have given rise to calls for greater transparency in the training data used to build the algorithms. These include proposals such as "datasheets" for training datasets \citep{gebru2018datasheets}, model cards \citep{mitchell2019model}, and robust algorithmic auditing both by citizens and governments \citep{raji2019actionable}. 

But who audits the auditor? In this paper we show that poorly defined standards of label quality can lead to spurious results in algorithmic audits. The challenge of quality control in machine learning datasets is widely known \citep{paullada2020data}. Our work highlights that even in "small data" settings like algorithmic audits, label quality can distort results. 

These pose new policy challenges for the deployment, auditing, and governance of AI systems, particularly in high-stakes settings. There are no universal benchmarks for label quality. In the context of algorithmic auditing, label quality is both a normative and economic question. Whose labels are better? What is an acceptable label quality? How much should we spend on creating high-quality labels for benchmarks? These are all highly practical and relevant questions about auditing AI algorithms in the real world. 

In this paper we use a new voice dataset collected in India to demonstrate that fidelity of  "ground truth data" can have serious implications on the results of algorithmic audits. Our dataset is collected from low-resourced settings in India. The Indian context is interesting for three reasons: technology adoption has grown exponentially in the last decade with over 740 million internet users and this number is projected to grow . The adoption of AI enabled applications in healthcare and education is particularly promising in many low-resource settings in India. Furthermore, India possesses a staggering diversity of languages, socioeconomic backgrounds which makes the design and governance of AI enabled services particularly challenging.  

\section{Related Work}
\label{work}
Algorithmic audits have become a popular strategy to highlight scenarios where AI applications can be biased against specific sub-populations \citep{guszcza2018audit}. For example, \citet{koenecke2020racial} conducted an audit of state-of-the-art, "off the shelf" automated speech recognition (ASR) systems developed by Amazon, Apple, Google, IBM, and Microsoft. They found that all five ASR systems exhibit significantly lower accuracy for Black English speakers (word error rate = 0.35) than for White English speakers (word error rate = 0.19). There is promise that such audits can influence AI development and governance. For example, \citet{raji2019actionable} show that an algorithmic audit of gender and skin type performance in commercially available facial recognition models led to developers significantly improving the accuracy of these models within a year.

Our work is closely related to \citet{koenecke2020racial}. We set out to conduct an audit of commercially available ASR systems from three primary cloud vendors in India (Amazon, Google, and Microsoft). However we ask a different, more "upstream" data quality question: how does the quality of {\it human} transcriptions affect the audit of ASR systems? This is a critical question for AI governance because widely accepted benchmarks on label quality do not exist, especially for audit of AI systems in developing countries. As far as we know, this has not been highlighted in the literature. 

%They report a word- error rate.   . Szymanski et al. (2020) use an independent conversational dataset to argue that WER on most ASR systems is significantly higher than reported values. Feng et al. (2021) use a Dutch voice corpus to      . 

While the focus of this paper is the need for clear label quality benchmarks needed for algorithmic audits, similar concerns about label quality have been recently discussed in the context of training data for ML algorithms. \citet{tsipras2020imagenet} use human studies to show that specific design choices in image annotation pipelines can lead to significant misalignment between a benchmark (like ImageNet) and the "ground truth" it serves as a proxy for. 

%Finally, our work is inspired by the rich body of recent work emphasizing the need for better data quality. Sambasivasan et al. (2021) highlight how undervaluing data quality leads to negative downstream effects of AI/ML ("data cascades") in high-stakes applications. Kshirsagar et al. (2021) emphasize how efforts to build "AI for Good" needs to explicitly confront the data quality challenge. For instance, one of their recommendations is: "In several socially important domains, labels suffer from subjective annotation. Such situation should be identified upfront to avoid introducing inconsistencies in the modeling pipeline." 

In this paper we empirically demonstrate how the issue of label quality can pose serious challenges in not just the development of new AI systems but auditing commercially available AI systems. This is a first order priority in AI governance that the data-centric AI movement needs to address.

\section{Materials and Methods}
\label{methods}

\subsection{Data}
\label{data}
We collected a novel dataset of conversational speech in India from 112 high school and college students. Our sample includes students from rural and urban areas and all students speak English as a second language. Thus this data is ideal to evaluate if widely deployed ASR systems in India perform poorly compared to global benchmarks and if there is significant regional disparities.

\begin{figure}[!h]
\label{fig1}
\centering
\includegraphics[height = 2.2 in, width = 4 in]{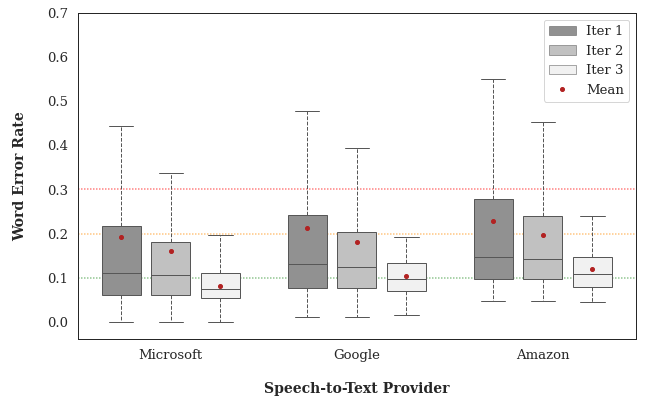}
\caption{Word Error Rate (WER) for three ASR providers (Microsoft, Google, and Amazon). We show the WER for the three iterations of data cleaning described in Section 3.1}
\end{figure}

We hired freelance transcribers (with an educational background similar to data annotators hired by AI labelling firms in India). These transcribers had prior experience in voice-to-text transcription. The transcriptions produced underwent three rounds of reviews (``iterations"):
\paragraph{Iteration 1:}This consists of the raw transcriptions we received from the hired transcribers without any edits. %The cost of this step was \todo{ADD COST}
\paragraph{Iteration 2:}This version consists of the research team at EquiTech Futures correcting instances where transcribers forced grammatical corrections (like paraphrasing and contracting sentences). The transcript after this iteration roughly matched the audio clips. %The cost of this step was %\todo{ADD COST}
\paragraph{Iteration 3:}This version is where all missing minute details were captured by a second review by the EquiTech Futures team. We added any missing filler words (eg: "so", "umm", "err"). We captured all the incomplete uttered words i.e. when the utterance was interrupted in the middle to form a related word (for eg: "the equation required simply... simplification"). %The cost of this step was \todo{ADD COST}

\subsection{Results}
\label{results}

We begin by assessing the performance of off-the-shelf ASR 
systems by primary cloud service providers (Amazon, Google, Microsoft)
using the word error rate (WER) defined as:
\begin{equation}
    \mathrm{WER} = \frac{S+D+I}{N}
\end{equation}
where $S/N$, $D/N$, and $I/N$ are the substitution, deletion, and insertion error rates respectively between the machine and human transcription. $N$ is the total
number of words in the human transcription. 

Figure 1 shows that for all three providers the mean WER decreases by $\sim10$\% between the raw transcript and the final iteration. In fact, after rigorous cleaning of the transcripts the WER for all providers is close to the industry recommended benchmark of WER$=0.1$.

\begin{figure}[!htb]
\centering
\includegraphics[height = 3.5 in, width = 5 in]{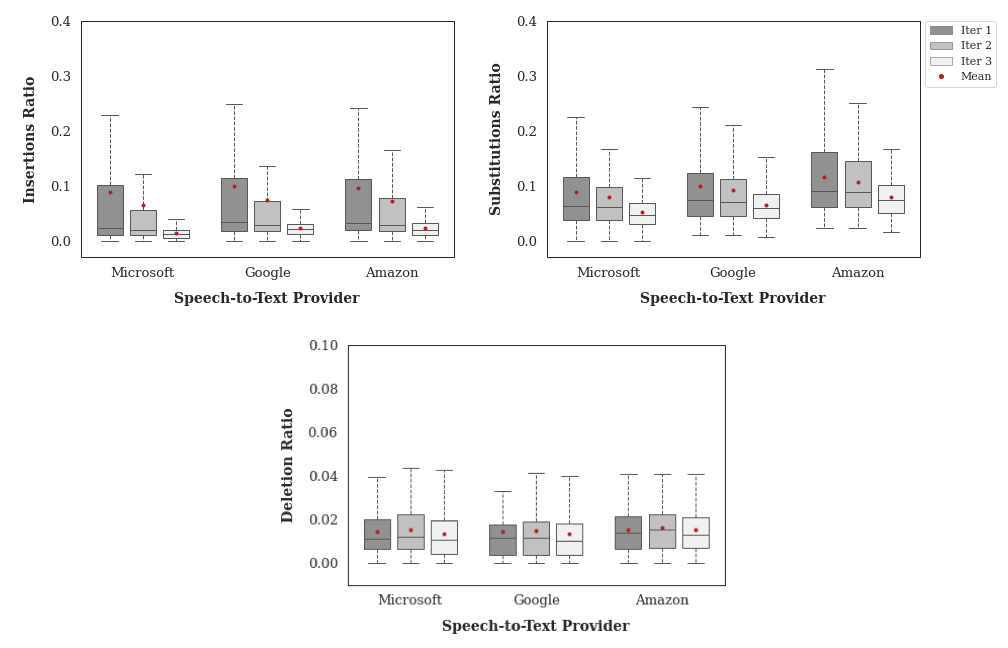}
\caption{Insertion ratio, substitution ratio, and deletion ratio for the three ASR providers disaggregated by three iterations of data cleaning.} 
\end{figure}

In Figure 2 we plot the disaggregated insertion, substitution, and deletion ratios between different iterations. Evidently, the human transcriptions do not significantly remove words in the original audio. The WER is dominated by transcribers making substitutions and inserting new words during the transcription process. 

Finally, in Figure 3 we investigate a hypothetical audit for ASR systems for rural vs urban populations for different iterations of the transcriptions. Transcriptions that have not been through detailed cross-checks would have revealed a stark but spurious difference in performance of ASR systems between rural and urban technology users. These differences disappear after correcting for the quality of ground-truth data. 

An interesting feature of our hypothetical audit is the improvement of ASR performance only for rural populations after subsequent data cleaning. This is because the rural English speakers in our sample have more grammatically incorrect sentences (eg: misplaced articles, incorrect prepositions, incomplete sentences). The human transcribers (who are urban) automatically correct these grammatical errors (even after being instructed not to) but the ASR transcribes the text faithfully. This leads to a spurious reduction in ASR performance driven by human annotation errors.

%We use our novel dataset to conduct a series of experiments to explore how errors in ASR transcriptions cascade to downstream NLP tasks like text summarization or question answering tasks. Our unique dataset allows us to measure the sensitivity of downstream ML tasks to errors in the ASR transcriptions. 

\begin{figure}[h]
\centering
\includegraphics[height = 2 in, width = 3 in]{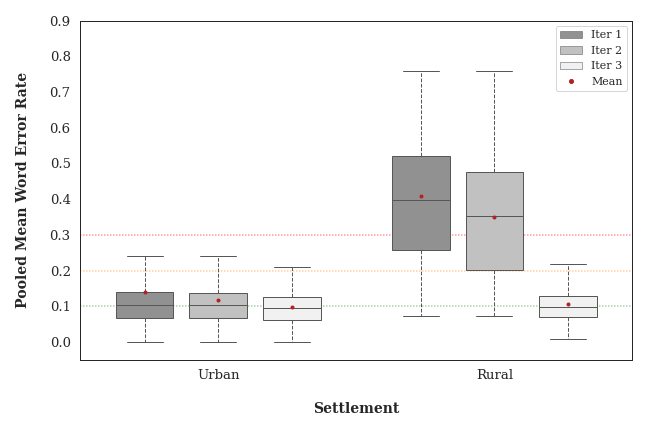}
\caption{Comparison of mean WER between urban and rural sub-samples in our dataset after different iterations of data cleaning described in Section 3.1.}
\end{figure}

\section{Discussion}
\label{discussion}
The initial goal of this project was to replicate results by \citet{koenecke2020racial} for the South Asian context by creating a similar audit for widely deployed ASR systems in India. We were especially interested in looking at disparities between rural and urban English speakers and low and high-income English speakers (since accents and fluency can vary significantly between these groups). During this process we discovered significant quality challenges with the transcribed "ground truth" data by human data labelers representative of data annotators typically employed in India. Our goal in writing this paper was to demonstrate empirically how algorithmic audits can suffer from challenges of data quality similar to the challenge confronted by algorithm developers. We hope our work highlights the need for widely accepted, consensus-driven standardized protocols for label quality. Such protocols are critical for AI governance globally.

\begin{ack}
We thank Bhasi Nair, Adam Bear, and Krittika Bhattacharjee for detailed comments on an earlier draft of this paper. Joyeeta Dey and Payal Goghare provided critical support in data collection and overseeing  data annotation. AM is supported by the Kevin Xu Initiative on Science, Technology, and Public Policy at the Harris School of Public Policy, University of Chicago. YG is supported by EquiTech Futures. 

%More information about this disclosure can be found at: \url{https://neurips.cc/Conferences/2021/PaperInformation/FundingDisclosure}.

\end{ack}

%References follow the acknowledgments. Use unnumbered first-level heading for
%the references. Any choice of citation style is acceptable as long as you are
%consistent. It is permissible to reduce the font size to \verb+small+ (9 point)
%when listing the references.
%Note that the Reference section does not count towards the page limit.
\medskip

%Imports the bibliography file "citations.bib"
\bibliography{citations}

\end{document}